\documentclass{article}
\usepackage{spconf,amsmath,graphicx,hyperref}
\usepackage{booktabs}   
\usepackage{multirow}   
\usepackage{siunitx}    
\usepackage{pgfplots}
\usepgfplotslibrary{polar} 
\pgfplotsset{compat=1.18} 
\usepackage{subcaption}
\usepackage{xcolor} 
\usepackage[table]{xcolor} 
\definecolor{human_superior}{gray}{0.9} 
\usepackage{tabularx}
\definecolor{softred}{RGB}{228, 26, 28}
\definecolor{softgreen}{RGB}{77, 175, 74}
\definecolor{userpurple}{RGB}{152, 118, 224}
\definecolor{softgrey}{RGB}{153, 153, 153}
\definecolor{softblue}{RGB}{55, 126, 184}
\definecolor{softorange}{RGB}{253, 174, 97}
\definecolor{taskbackground_gray}{gray}{0.95}

\title{THE MUSE BENCHMARK: PROBING MUSIC PERCEPTION AND AUDITORY RELATIONAL REASONING IN AUDIO LLMS}
\threeauthors
 {Brandon James Carone}
	{New York University\\
	Department of Psychology\\
    Music and Audio Research Lab\\
	New York, NY 10012, USA}
 {Iran R. Roman}
	{Queen Mary University of London\\
    School of EECS\\
    Centre for Multimodal AI\\   
        London, England, UK}
{Pablo Ripollés}
	{New York University\\
	Department of Psychology\\
    Music and Audio Research Lab\\
	New York, NY 10012, USA}

\begin{document}
\ninept
\maketitle
\begin{abstract}
Multimodal Large Language Models (MLLMs) have demonstrated capabilities in audio understanding, but current evaluations 
may obscure fundamental weaknesses in relational reasoning. We introduce the Music Understanding and Structural Evaluation (MUSE) Benchmark, an open-source resource with 10 tasks designed to probe fundamental music perception skills. We evaluate four SOTA models (Gemini Pro and Flash, Qwen2.5-Omni, and Audio-Flamingo 3) against a large human baseline (N=200). Our results reveal a wide variance in SOTA capabilities and a persistent gap with human experts. While Gemini Pro succeeds on basic perception, Qwen and Audio Flamingo 3 perform at or near chance, exposing severe perceptual deficits. Furthermore, we find Chain-of-Thought (CoT) prompting provides inconsistent, often detrimental results. Our work provides a critical tool for evaluating invariant musical representations and driving development of more robust AI systems.
\end{abstract}
\vspace{-4pt}
\begin{keywords}
Benchmarks, Music Understanding, Multimodal LLMs, Human-Computer Comparison
\end{keywords}
\section{Introduction}
\label{sec:intro}
Recent advances in large multimodal models have extended the foundation-model paradigm to audio. Systems such as Google’s Gemini 2.5 \cite{Comanici2025}, Alibaba’s Qwen2.5-Omni \cite{Xu2025}, and NVIDIA’s Audio Flamingo 3 \cite{Goel2025} demonstrate competitive performance across  audio benchmarks covering speech recognition, tagging/captioning, and in-the-wild Question Answering (e.g., AIR-Bench~\cite{Yang2024}; MMAR~\cite{Ma2025}; MMAU~\cite{Sakshi2024}; MMAU-Pro~\cite{Kumar2025}). Yet these evaluations largely probe surface-level classification rather than deeper perceptual understanding~\cite{carone2025evaluating}. We argue that current benchmarks do not test abstract, relational reasoning in music, such as pitch-invariant recognition of a melody under transposition, or perception of melodic contour and chord harmonic function. These abilities are fundamental to human hearing and are documented across expertise levels~\cite{Krumhansl1990, Krumhansl2004, Krumhansl2010, Vuust2014, Kim2022, Bonetti2024}. While benchmarks on tasks such as genre identification or descriptive captioning may indicate that models ``understand music", 
, they may succeed by learning surface co-occurrences (e.g., timbre or tempo cues) rather than the relations that constitute musical structure.

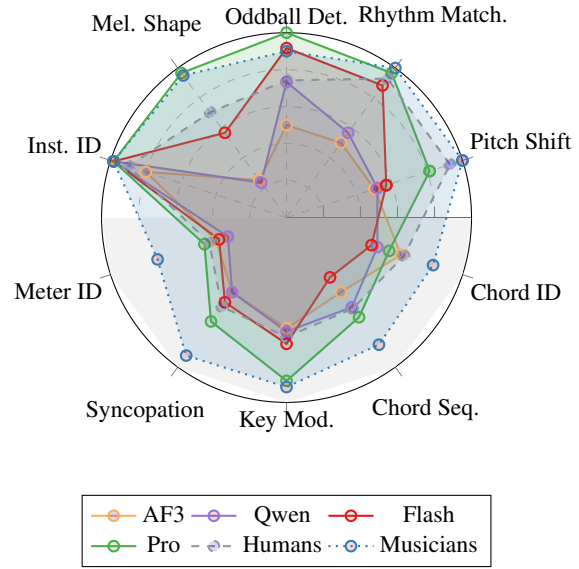
\begin{figure}[t!]
\centering
\begin{tikzpicture}

\begin{polaraxis}[
    width=6.5cm,
    height=6.5cm,
    legend style={
        at={(0.5, -0.25)},
        anchor=north,
        legend columns=3,
        font=\small,
    },
    xtick={18, 54, 90, 126, 162, 198, 234, 270, 306, 342},
    xticklabels={
        Pitch Shift,
        Rhythm Match.,
        Oddball Det.,
        Mel. Shape,
        Inst. ID,
        Meter ID,
        Syncopation,
        Key Mod.,
        Chord Seq.,
        Chord ID
    },
    xticklabel style={font=\small}, 
    ymin=0, ymax=100,
    ytick={20, 40, 60, 80}, 
    yticklabels={}, 
    yticklabel style={anchor=south east, font=\small, color=gray},
    grid=both,
    major grid style={dashed, color=gray!50},
]

\addplot+[taskbackground_gray, fill=taskbackground_gray, draw=none, mark=none, forget plot]
    coordinates {(180, 100) (198, 100) (234, 100) (270, 100) (306, 100) (342, 100) (360, 100)} \closedcycle;

\addplot+[color=softorange, mark=*, thick, fill=softorange, fill opacity=0.15] coordinates { (18, 50.00) (54, 50.00) (90, 50.00) (126, 25.00) (162, 80.00) (198, 40.00) (234, 50.00) (270, 60.00) (306, 50.00) (342, 65.00) (18, 50.00) };
\addplot+[color=userpurple, mark=*, thick, fill=userpurple, fill opacity=0.15] coordinates { (18, 51.67) (54, 56.67) (90, 73.33) (126, 23.33) (162, 98.33) (198, 33.33) (234, 50.00) (270, 61.67) (306, 60.00) (342, 51.67) (18, 51.67) };
\addplot+[color=softred, mark=*, thick, fill=softred, fill opacity=0.15] coordinates { (18, 56.67) (54, 88.33) (90, 91.67) (126, 56.67) (162, 98.33) (198, 38.33) (234, 56.67) (270, 68.33) (306, 40.00) (342, 48.33) (18, 56.67) };
\addplot+[color=softgreen, mark=*, thick, fill=softgreen, fill opacity=0.15] coordinates { (18, 81.36) (54, 96.67) (90, 100.00) (126, 96.67) (162, 98.33) (198, 46.67) (234, 69.49) (270, 88.33) (306, 66.67) (342, 58.33) (18, 81.36) };

\addplot+[color=softgrey, mark=*, thick, dashed, fill=softgrey, fill opacity=0.15] coordinates { (18, 92.90) (54, 92.90) (90, 74.20) (126, 70.30) (162, 89.90) (198, 43.90) (234, 59.60) (270, 64.60) (306, 60.90) (342, 66.80) (18, 92.90) };
\addplot+[color=softblue, mark=*, thick, dotted, fill=softblue, fill opacity=0.15] coordinates { (18, 100.00) (54, 100.00) (90, 90.00) (126, 95.00) (162, 98.30) (198, 73.30) (234, 92.30) (270, 91.70) (306, 85.00) (342, 83.30) (18, 100.00) };

\legend{AF3, Qwen, Flash, Pro, Humans, Musicians}

\node[font=\small] at (axis cs:90, 125) {\textbf{Beginner Tasks}};
\node[font=\small] at (axis cs:270, 125) {\textbf{Advanced Tasks}};

\end{polaraxis}
\end{tikzpicture}
\caption{SOTA model comparison on the MUSE benchmark. Models shown with solid lines. Humans shown with dashed and dotted lines.}
\vspace{-12pt}
\label{fig:radar_chart_comparison}
\end{figure}

\begin{table*}[t]
\vspace{-6pt}
\caption{Overview of the 10 tasks in The MUSE Benchmark. All tasks contained 20 trials each.}
\vspace{-10pt}
\label{tab:task_overview}
\centering
\footnotesize
{ 
\renewcommand{\arraystretch}{1.5} 
\begin{tabularx}{\textwidth}{l l >{\raggedright\arraybackslash}X l >{\raggedright\arraybackslash}X}
\toprule
\textbf{Tier} & \textbf{Task Name} & \textbf{Technical Description} & \textbf{Input} & \textbf{Output Choices} \\
\midrule

\multirow{5}{*}{\textbf{Beginner}}
& Pitch Shift Detection & Detect whether a melody is pitch shifted. & Two audio clips & Same/Different \\
& Rhythm Matching & Determine if two rhythmic sequences match. & Two audio clips & Same/Different \\
& Oddball Detection & Detect out-of-key deviants in a melody. & Two audio clips & Same/Contains Oddball \\
& Instrument ID & Identify the instrument. & One audio clip & Piano/Guitar/Bass/Drums \\
& Melody Shape ID & Identify the overall melodic shape. & One audio clip & Ascending/Descending/Arch/Inv. Arch \\
\cmidrule(l){2-5} 

\multirow{5}{*}{\textbf{Advanced}}
& Chord Identification & Identify chord quality (major or minor). & One audio clip & Major/Minor \\
& Syncopation  & Determine which rhythm is more syncopated. & Two audio clips & Pattern 1/Pattern 2 \\
& Key Modulation & Detect if a change of key occurs. & One audio clip & Modulation/No Modulation \\
& Chord Seq. Matching & Determine if two chord sequences match. & Two audio clips & Same/Different \\
& Meter Identification & Identify the underlying grouping of beats. & One audio clip & Groups of 3/Groups of 4/Groups of 5 \\

\bottomrule
\end{tabularx}
} 
\vspace{-15pt}
\end{table*}

\section{The MUSE Benchmark}
\label{sec:benchmark}

The Music Understanding and Structural Evaluation (MUSE) Benchmark comprises 10 tasks divided into “Beginner” and “Advanced” tiers. The design is grounded in music cognition research to systematically probe for abstract, relational reasoning in audio models \cite{Dowling1971, Dowling1978, Deutsch1969, Deutsch1972}. To validate the benchmark's design, we also collected data from a large human sample. Table \ref{tab:task_overview} details tasks. \footnote{Full task descriptions and stimuli are available on \url{https://github.com/brandoncarone/MUSE_music_benchmark} and \url{https://airtable.com/appQCPXVEeadwacMP/shrHV0OjuwxYBzJ78}}.

\subsection{Beginner Tasks: Core Perception \& Invariance}
The five Beginner tasks target fundamental aspects of music perception, robust even in non-musicians \cite{Dowling1971, Trainor2010, Halpern2010}, and test a model’s ability to learn core auditory invariances. Instrument Identification assesses the ability to classify instruments based on their unique timbral qualities~\cite{Schellenberg2005, Giordano2010, McAdams2013}. Melody Shape Identification probes the recognition of a melody's overall shape (e.g., ascending/descending), a key aspect of melodic perception~\cite{Huron1996, Goldstein2024}. Oddball Detection evaluates sensitivity to tonal hierarchies by requiring the detection of out-of-key notes based on harmonic context~\cite{Krumhansl1979, Bharucha1983, Tervaniemi2003}. Rhythm Matching tests the processing of rhythmic sequences, a skill engaging both auditory and motor systems~\cite{Thaut2014, Grahn2007, Toiviainen2020}. Finally, Pitch Shift Detection assesses pitch-invariant melody recognition across transpositions, a process reliant on relative pitch and melodic contour \cite{Dowling1978, Deutsch1969, Deutsch1972}.

\subsection{Advanced Tasks: Music-Theoretic Skills}
These tasks target skills requiring formal musical training, 
demanding explicit knowledge of music-theoretic constructs and tracking of functional relationships over time \cite{Krumhansl1990, Temperley2001}. Three tasks probe harmonic understanding. Chord Identification requires distinguishing major and minor chords \cite{Bidelman2011, MacLean2024}. Chord Sequence Matching tests the recognition of functional harmonic patterns across different musical styles \cite{Wall2020}. Lastly, Key Modulation Detection evaluates a model’s capacity to represent tonal hierarchies and track changes of tonal center within an excerpt. Two tasks assess hierarchical rhythmic processing: Meter Identification requires inferring the underlying cycle of strong and weak beats from a surface rhythm \cite{Kondoh2021}, while Syncopation Comparison requires identifying off-beat accents by comparing a rhythm against an internalized meter \cite{Large2015, Fiorin2024}.

\section{Methods}
\label{sec:methods}

\subsection{Stimuli Creation}
    We composed and recorded 200 musical stimuli (mean length = 14.1sec, min = 3sec, max = 46sec) using Logic Pro X, an Apollo Twin X audio interface, Yamaha HS8 monitors, and a 2021 16” Macbook M1 Pro laptop. For the guitar recordings, both a PRS McCarty Hollowbody II and a Schecter Solo-6 were recorded using the Neural DSP Tim Henson Archetype and Cory Wong Archetype plugins. A Fender Squier Classic Vibe '60s Mustang Bass was played through the Neural DSP Cory Wong Archetype plugin for the bass recordings. The piano recordings were made using the Arturia KeyLab Essential Mk3 49-Key MIDI Keyboard Controller and the Analog Lab V plugin. Finally, a Roland TD-17 Electronic Drum Kit and the Superior Drummer 3 plugin were used for the drum recordings.

\begin{table*}[t]
\vspace{-6pt}
\caption{Accuracy 
on ten music perception tasks, 
separated by prompting condition. Five beginner tasks assess fundamental perceptual abilities: Instrument ID, Melody Shape ID, Oddball Detection, Rhythm Matching, and Pitch Shift Detection. Five Advanced tasks test 
skills requiring formal musical training: Chord ID, Key Modulation, Chord Sequence Matching, Syncopation Comparison, and Meter ID. Comparison tasks that require the processing of two audio files to answer a single question have a star (*) next to the name. Refer to Sections 2.1 and 2.2 for greater detail. The best-performing model per task/condition is shown in \textbf{bold} (second-best \underline{underlined}) and chance level is listed at the bottom. \textbf{Human \& Musician scores with a gray background indicate performance superior to the best model.} } 
\vspace{-7pt}
\label{tab:results_all_tasks_restructured}
\centering
\scriptsize 
\setlength{\tabcolsep}{4pt} 
\begin{tabular}{l l S[table-format=3.2] S[table-format=3.2] S[table-format=3.2] S[table-format=3.2] S[table-format=3.2] | S[table-format=3.2] S[table-format=3.2] S[table-format=3.2] S[table-format=3.2] S[table-format=3.2]}
\toprule
& & \multicolumn{5}{c}{\textbf{Beginner Tasks}} & \multicolumn{5}{c}{\textbf{Advanced Tasks}} \\
\cmidrule(lr){3-7} \cmidrule(lr){8-12}
\textbf{Strategy} & \textbf{Model} & {Inst. ID} & {Mel. Shape} & {Oddball Det.*} & {Rhythm Match.*} & {Pitch Shift*} & {Chord ID}& {Chord Seq. Match.*}  & {Key Mod.} & {Syncopation*} & {Meter ID} \\
\midrule

\multirow{4}{*}{Standalone}
& AF3   & {80.00}          & {25.00}          & {50.00}          & {50.00}          & {50.00}          & \textbf{65.00} & {50.00}          & {60.00}          & {50.00}          & \underline{40.00} \\
& Qwen  & \textbf{98.33} & {23.33}          & {73.33}          & {56.67}          & {51.67}          & {51.67}          & \underline{60.00}          & {61.67}     & {50.00}     & {33.33} \\
& Flash & \textbf{98.33} & \underline{56.67} & \underline{91.67} & \underline{88.33} & \underline{56.67} & {48.33}         & {40.00}  & \underline{68.33}     & \underline{56.67}     & {38.33} \\
& Pro   & \textbf{98.33} & \textbf{96.67} & \textbf{100.00} & \textbf{96.67} & \textbf{81.36} & \underline{58.33} & \textbf{66.67} & \textbf{88.33} & \textbf{69.49}  & \textbf{46.67} \\

\midrule
\multirow{4}{*}{CoT}
& AF3   & {70.00}          & {25.00}          & {50.00}          & {50.00}          & {50.00}          & \underline{50.00} & {50.00}          & {50.00}          & {50.00}          & \underline{40.00} \\
& Qwen  & \textbf{98.33} & {18.33}          & {70.00}          & {50.00}          & {58.33}          & {48.33}   & \textbf{50.00}        & {48.33}          & \underline{50.00} & {35.00} \\
& Flash & \underline{91.67} & \underline{46.67} & \underline{85.00} & \underline{63.33} & \underline{86.67} & {43.33}          & \underline{48.33}         & \underline{58.33} & {43.33}  & {35.00} \\
& Pro   & \textbf{98.33} & \textbf{96.67} & \textbf{100.00} & \textbf{88.33} & \textbf{98.33} & \textbf{56.67} & {46.67} & \textbf{81.67}    & \textbf{61.67}       & \textbf{50.00} \\
\midrule
& Humans & {89.90} & {70.30} & {74.20} & \cellcolor{human_superior}{92.90} & \cellcolor{human_superior}{92.90} & \cellcolor{human_superior}{66.80} & {60.90} & {64.60} & {59.60} & {43.90} \\
& Musicians  & {98.30} & {95.00} & {90.00} & \cellcolor{human_superior}{100.00} & \cellcolor{human_superior}{100.00} & \cellcolor{human_superior}{83.30} & \cellcolor{human_superior}{85.00} & \cellcolor{human_superior}{91.70} & \cellcolor{human_superior}{92.30} & \cellcolor{human_superior}{73.30} \\
\midrule
& Chance & {25.00} & {25.00} & {50.00} & {50.00} & {50.00} & {50.00} & {50.00} & {50.00} & {50.00} & {33.00} \\
\bottomrule
\end{tabular}
\vspace{-12pt}
\end{table*}

\subsection{Model Evaluation}
We implemented custom inference scripts to standardize prompt delivery and response recording for four SOTA models: Audio Flamingo 3, Qwen2.5-Omni, Gemini 2.5 Flash, and Gemini 2.5 Pro. For tasks requiring the comparison of two musical stimuli, we accommodated each model's specific input constraints. While the Qwen and Gemini models allow for multiple audio files to be processed in one turn, Audio Flamingo 3 can only process one excerpt at a time. For this model, the two stimuli were concatenated into a single audio file, separated by spoken verbal cues (“Here is the first excerpt,” “Here is the second excerpt”) and brief silences (1-2secs).

We evaluated all models in two distinct prompting conditions: \textbf{Standalone:} Mirrors the human experiment. To ensure models could maintain memory across trials—analogous to a human's ability to remember task instructions—we utilized the models' chat modes, which are optimized for multi-turn, stateful interactions \cite{Lampinen2022, Lingfan2025}. System instructions and few-shot examples provided to the models were identical to those given to human participants. \textbf{Chain-of-Thought (CoT):} We augmented the prompts to instruct the models on a multi-step analytical process (e.g., abstracting harmonic function, comparing rhythmic patterns). Few-shot examples provide a complete in-context demonstration of this process, with the model-side response explicitly articulating its reasoning for each step before providing the final answer.

A necessary exception was made for Audio Flamingo 3. Preliminary testing revealed that it failed to follow instructions reliably with chat history maintained and with few-shot examples. In these conditions it effectively performed at chance level. Therefore, Audio Flamingo 3 was evaluated without chat history and examples, using a combined system and per-trial prompt. See Table A on the Github repo for a summary of prompting strategies.

To get a stable and reliable measure of each model’s performance, we accounted for the stochastic nature of LLMs. Each task script was run three times with different random seeds, and the resulting accuracies were averaged per task. This resulted 240 runs total (4 models × 10 tasks × 2 prompting strategies  × 3 seeds), allowing us to account for model nondeterminism. All inference scripts were uniformly structured to include:
\textbf{1) System Instructions} specific to each task, provided before any interaction. See the scripts in the Github repo for system instructions.    \textbf{2) In-context Few-shot Learning} \cite{Brown2020, Ouyang2022, Wei2022}, where models are conditioned on several task demonstrations provided directly in the prompt at inference time, without any gradient updates. One example was given for every possible answer choice, except for the Audio Flamingo 3 condition. \textbf{3) Standardized Audio Presentation} and deterministic response formatting (e.g., “Yes, these are the same exact melody.”).    \textbf{4) Systematic Data Logging} of all outputs for later analysis.

\subsection{Human Data Collection}
We also collected human data from 234 online participants. To ensure data quality, we excluded 34 participants who failed an in-experiment headphone check \cite{woods2017headphone}, resulting in a final sample of 200 participants (105 males, 89 females, 6 non-binary; mean age = 38.76, SD = 12.79) recruited via Prolific  and New York University’s student population. The experiment was implemented in PsychoPy \cite{Peirce2019} and hosted on Pavlovia. To assess musical expertise, participants completed the Goldsmiths Musical Sophistication Index (Gold-MSI; \cite{Mullensiefen2014}. Using the score norms from the Musical Training scale, we segmented our sample into an 'Overall' group (N=200) and an 'Expert Musician' subgroup (N=6), defined as those scoring in the 90th percentile or higher. To mitigate fatigue over the long experiment, the benchmark was divided into two halves, with participants randomly assigned to one; for final analysis, accuracy was calculated by pooling the number of correct responses across both groups for each task. The order of tasks and stimuli was randomized to prevent order effects. Prior to each of the 10 tasks, participants received detailed instructions and few-shot examples for every possible answer choice (e.g., two examples for binary tasks, four for 4-alternate forced choice tasks), which matched those of the models for the Standalone condition.\footnote{Full human data available here: \url{https://osf.io/pvrd7/?view_only=3c3ac357272e43a08a201698fe6bd9c9}}

\section{Results and Discussion}
\label{sec:results}
Figure \ref{fig:radar_chart_comparison} and Table \ref{tab:results_all_tasks_restructured} present the benchmark's results. Overall, models in the Standalone condition matched or outperformed their CoT counterparts. We report and discuss these results in detail below. 

\subsection{The Human-Machine Gap in Music Reasoning}
Human listeners, especially expert musicians, consistently outperformed most models on tasks requiring abstract reasoning  (i.e., Melody Shape ID, Pitch Shift Detection) and those requiring knowledge of music theory (i.e., all tasks in the Advanced tasks). While top models were competitive on classification-style tasks like Instrument Identification (Gemini Pro and Qwen achieve 98.33\% accuracy, matching the 98.30\% of experts), a significant gap emerges in relational tasks. For example, expert musicians achieved perfect accuracy (100\%) on Pitch Shift Detection, whereas the best model (Gemini Pro) required CoT prompting to reach a similar level (81.36\% in Standalone Condition). This gap is even more pronounced in the advanced tasks, which require one to extract rhythmic and pitch information, maintain them in memory, and establish physical relations between auditory objects. Human musical experts consistently outperformed all models on complex harmonic and rhythmic judgments, achieving 85.00\% on Chord Sequence Matching and 91.70\% on Key Modulation Detection, compared to Gemini Pro’s scores of 66.67\% and 88.33\%, respectively. The disparity is particularly large in Meter Identification, where human music experts (73.30\%) substantially outperformed the best model (Gemini Pro, 46.67\%).
Interestingly, Melody Shape Identification revealed high variance among models rather than a simple human-machine gap. Human music experts scored at 95.00\%, with Gemini Pro performing similarly (96.67\%). However, other SOTA models failed dramatically on this same task.

\vspace{-6pt}
\subsection{Critical Failures Reveal Limits of SOTA Models}
Our benchmark uncovers not just performance gaps but critical failures in some SOTA models. Most notably, Qwen's accuracy on Melody Shape Identification (23.33\%) is around the 25\% chance level, indicating a fundamental failure to process relative pitch direction. Audio Flamingo 3 exhibits a more widespread lack of competence, performing at or just above chance on nearly all of the 10 tasks. 
While Gemini Pro was the strongest model overall, its performance profile reveals a clear hierarchy of difficulty. It achieved perfect (100\% on Oddball Detection) or near-perfect (96.67\% on Rhythm Matching) scores on beginner tasks with clear acoustic cues. However, its accuracy declined on advanced tasks requiring more abstract, relational reasoning, such as Chord Sequence Matching (66.67\%) and Meter Identification (46.67\%). 

\vspace{-6pt}
\subsection{CoT Prompting is Unreliable and Inconsistent}
The application of CoT prompting yielded inconsistent and often detrimental results, revealing its unreliability for complex audio reasoning. CoT only produced a dramatic improvement in one case, boosting Gemini Pro’s Pitch Shift Detection accuracy from 81.36\% to a near-human 98.33\%. More frequently, CoT either had a negligible effect or actively harmed performance. CoT degraded Gemini Pro's accuracy on Rhythm Matching (from 96.67\% to 88.33\%) and Syncopation Comparison (from 69.49\% to 61.67\%). Similarly, it worsened Qwen's already below-chance score on Melody Shape Identification (from 23.33\% to 18.33\%). The inconsistent effects of CoT—sometimes boosting, other times harming performance—show that step-by-step textual reasoning is not a reliable way to enhance models' non-linguistic perceptual skills

Analysis of Gemini Pro’s CoT logs reveals that the model often sounds correct while reasoning incorrectly. In Syncopation Comparison, its reasoning was directionally consistent in all 37 correct trials, but the precise off-beat counts were correct in only 4/37. For Chord Quality ID, it correctly identified the defining major or minor third in 34/60 trials; incorrect responses either asserted the opposite quality or offered vague qualitative language. In Chord Sequence Matching, the model showed a strong bias, correctly identifying progressions like I–V–vi–IV (19 times) and vi–IV–I–V (8 times) but never correctly identifying others (e.g., I–IV–V). Finally, for 30 modulation items in Key Modulation Detection, the model incorrectly asserted “no modulation” in 10 cases, and for the 27 trials it did describe, the mean absolute error was 3.04 scale degrees with only one exact match. Overall, while the CoT explanations sounded confident, they were not always truly dependable.

\vspace{-6pt}
\subsection{Post-hoc analyses: Comparing Model In-Context Learning to Human Learning}
To test whether models ``learn" from repeated examples as humans do through musical training, we compared human musical expertise with Gemini models' in-context learning by varying the number of few-shot examples ({0,1,2,4,8} or {0,1,3,6,9}, depending on response options) \cite{Brown2020}. We therefore use the number of shots as a proxy for this learning process to test whether models, like humans, consistently improve on complex musical tasks with greater exposure to few-shot examples. For the model analysis, we first pooled the results from Gemini Pro and Flash within each task and fit Generalized Linear Models (GLM) to estimate the effect of number of shots (for models) on task accuracy. For humans, we used a Generalized Linear Mixed-Effects Model (GLMM) to estimate the effect of musical training (Gold-MSI Training scale) on accuracy for each task. For all models, the primary effect size was the regression coefficient for the predictor of interest (Number of Shots or Musical Training), representing the change in the log-odds of a correct response for ever one-unit increase in the predictor.

We focused on the four tasks that showed the most dynamic performance changes for the models: Melody Shape ID, Key Modulation Detection, Chord Sequence Matching, and Syncopation Comparison (full results for all tasks are in Table B, GitHub repository). This analysis was limited to the Gemini models as they demonstrated the most accurate scores on the benchmark. We excluded conditions where models were already at ceiling (e.g., for Melody Shape ID we used Flash accuracy, as Gemini Pro was at ceiling). This focused approach allows for a clear comparison between the learning patterns of SOTA models and those of human participants.

\begin{figure}[t]
    \centering 

    \includegraphics[width=\columnwidth]{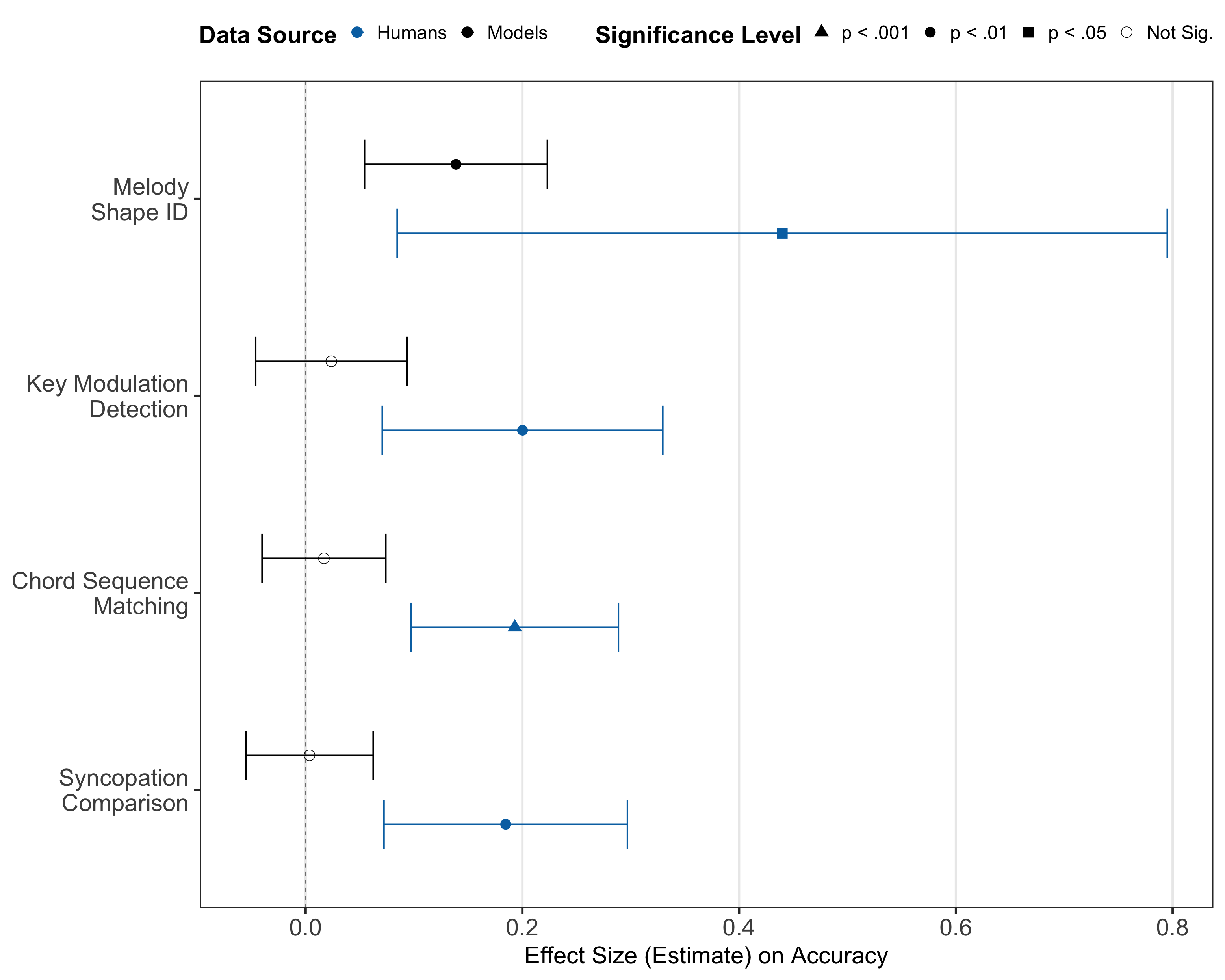}
    \vspace{-12pt}
    \caption{Relationship between number of shots provided and accuracy across Gemini Pro and Gemini Flash (black). The relationship between human accuracy and musical training is also shown (blue). Points represent the estimated effect size (log-odds ratio) from a GLM for the models, and a GLMER for the humans, for each task. Error bars indicate the 95\% confidence interval. Positive estimates mean greater shots or training correspond to higher accuracy. The shape of each point indicates the statistical significance of the effect.}   

    \label{fig:human_training_effect}
    \vspace{-8pt}
\end{figure}

The results, visualized in Figure \ref{fig:human_training_effect}, reveal that the effect of providing more in-context shots to the models was inconsistent and task-dependent. A significant positive effect was found for only one task: Melody Shape ID (p $<$ .001), and this was only true for Gemini Flash (we did not run the extra shot conditions for Pro since it was already at ceiling). This suggests that Gemini Flash may leverage more examples to improve performance on tasks that rely on recognizing clear, repeating perceptual patterns. However, for the three tasks requiring more abstract, music understanding, such as Key Modulation Detection, Chord Sequence Matching, and Syncopation Comparison, the number of shots had no statistically significant effect on model accuracy. 

In stark contrast, the results reveal a clear and cognitively plausible pattern for human performance. Musical training had a significant, positive effect on accuracy across all four tasks in our analysis. This confirms that for humans, dedicated training corresponds to the internalization of abstract rules that reliably improve performance on both foundational and advanced musical judgments.

This analysis demonstrates a fundamental divergence between human learning and the models' in-context learning on these tasks. While musical training in humans corresponds to the internalization of abstract rules that reliably improve performance, providing models with more examples is an unreliable proxy for such training. The models performance seems more dependent on their pre-trained capabilities, which are conditioned by a small number of shots but not consistently improved by more. This suggests that bridging the gap in music understanding between humans and machines may require fundamental changes in model training paradigms (perhaps mimicking the way humans learn music), rather than simply providing more in-context examples at inference time.

\vspace{-6pt}
\section{CONCLUSION}
\vspace{-6pt}
Our evaluation of SOTA models on the MUSE benchmark reveals a gap against human experts, particularly on tasks requiring abstract relational reasoning. While top models like Gemini Pro succeed on basic perception, their accuracy declines on advanced tasks involving harmony and meter. Other models fail at or below chance, indicating a shared lack of invariant musical representations. We also find that common prompting strategies are unreliable; CoT was often detrimental, and increasing few-shot examples did not produce consistent learning effects. In conclusion, the MUSE benchmark provides a critical diagnostic tool, revealing that current audio LLMs lack the invariant representations necessary for deep musical reasoning. Our results challenge the field to move beyond surface-level classification and motivate the development of foundation models that target genuine perceptual competence. Bridging the human-machine gap in music will likely require fundamental changes in model architecture and training paradigms, rather than simply scaling existing methods with more data or more complex prompts.



\bibliographystyle{IEEEbib}
\bibliography{refs}

\end{document}